\documentclass{article}

\usepackage{dsfont}
\usepackage{amssymb}
\usepackage{amstext}
\usepackage{amsmath}
\usepackage{amsfonts}
\usepackage{amsthm}

\usepackage{graphicx} 
\usepackage{subfigure}

\usepackage{natbib}

\usepackage{algorithm}
\usepackage{algorithmic}

\usepackage{hyperref}

\usepackage[accepted]{icml2013}

\icmltitlerunning{Physeter catodon localization by sparse coding}

\newcommand{\g}[1]{\mbox{\boldmath{$#1$}}}

\begin{document}

\twocolumn[
\icmltitle{Physeter catodon localization by sparse coding}

\icmlauthor{S\'{e}bastien PARIS}{sebastien.paris@lsis.org}
\icmladdress{DYNI team, LSIS CNRS UMR 7296, Aix-Marseille University}
\icmlauthor{Yann DOH}{yanndoh.m2@gmail.com}
\icmladdress{DYNI team, LSIS CNRS UMR 7296, Universit\'{e} Sud Toulon-Var}
\icmlauthor{Herv\'{e} GLOTIN}{glotin@univ-tln.fr}
\icmladdress{DYNI team, LSIS CNRS UMR 7296, Universit\'{e} Sud Toulon-Var}
\icmlauthor{Xanadu HALKIAS}{halkias@univ-tln.fr}
\icmladdress{DYNI team, LSIS CNRS UMR 7296, Universit\'{e} Sud Toulon-Var}
\icmlauthor{Joseph RAZIK}{razik@univ-tln.fr}
\icmladdress{DYNI team, LSIS CNRS UMR 7296, Universit\'{e} Sud Toulon-Var}

\icmlkeywords{spermwhale, clicks, whale watching, bag-of-features, sparse coding, machine learning, regression, particle filter}

\vskip 0.3in
]

\begin{abstract}
This paper presents a spermwhale' localization architecture using jointly a bag-of-features (BoF) approach and machine learning framework. BoF methods are known, especially in computer vision, to produce from a collection of local features a global representation invariant to principal signal transformations. Our idea is to regress supervisely from these local features two rough estimates of the distance and azimuth thanks to some datasets where both acoustic events and ground-truth position are now available. Furthermore, these estimates can feed a particle filter system in order to obtain a precise spermwhale' position even in mono-hydrophone configuration. Anti-collision system and whale watching are considered applications of this work.
\end{abstract}

\section{Introduction}
\label{Introduction}

Most of efficient cetacean localisation systems are based on the Time Delay Of Arrival (TDOA) estimation from detected\footnote{As click/whistles detector, matching filter is often prefered} animal's click/whistles signals \cite{nosalGT,fredericbenard_herveglotin_2009}. Long-base hydrophones'array is involving several fixed, efficient but expensive hydrophones \cite{3Dtracking} while short-base version is requiring a precise array's self-localization to deliver accurate results. Recently (see \cite{IFA_DCL}), based on Leroy's attenuation model versus frequencies \cite{Leroy}, a range estimator have been proposed. This approach is working on the detected most powerful pulse inside the click signal and is delivering a rough range' estimate robust to head orientation variation of the animal. Our purpose is to use i) these hydrophone' array measurements recorded in diversified sea conditions and ii) the associated ground-truth trajectories of spermwhale (obtained by precise TDAO and/or Dtag systems) to regress both position and azimuth of the animal from a third-party hydrophone\footnote{We assume that the velocity vector is colinear with the head's angle.} (typically onboard, standalone and cheap model).

We claim, as in computer-vision field, that BoF approach can be successfully applied to extract a global and invariant representation of click's signals. Basically, the pipeline of BoF approach is composed of three parts: i) a local features extractor, ii) a local feature encoder (given a dictionary pre-trained on data) and iii) a pooler aggregating local representations into a more robust global one. Several choice for encoding local patches have been developed in recent years: from hard-assignment to the closest dictionary basis (trained for example by $K$means algorithm) to a sparse local patch reconstruction (involving for example Orthognal Maching Pursuit (OMP) or LASSO algorithms).

\section{Global feature extraction by spare coding}

\subsection{Local patch extraction}

Let's denote by $\g{C}\triangleq\{\g{C}^j\}$, $j=1,\ldots,H$ the collection of detected clicks associated with the $j^{th}$ hydrophone of the array composed by $H$ hydrophones. Each matrix $\g{C}^j$ is defined by $\g{C}^j\triangleq\{\g{c}_i^j\}$, $i=1,\ldots,N^j$ where $\g{c}_i^j\in\mathds{R}^n$ is the $i^{th}$ click of the $j^{th}$ hydrophone. For our \textit{Bahamas2} dataset \cite{3Dtracking}, we choose typically $n=2000$ samples surrounding the detected click. The total number of available clicks is equal to $N=\sum\limits_{i=1}^{H}N^j$.

\begin{figure*}[!ht]
\begin{center}
\begin{tabular}{cc}
\includegraphics[height=5cm,width=7.5cm]{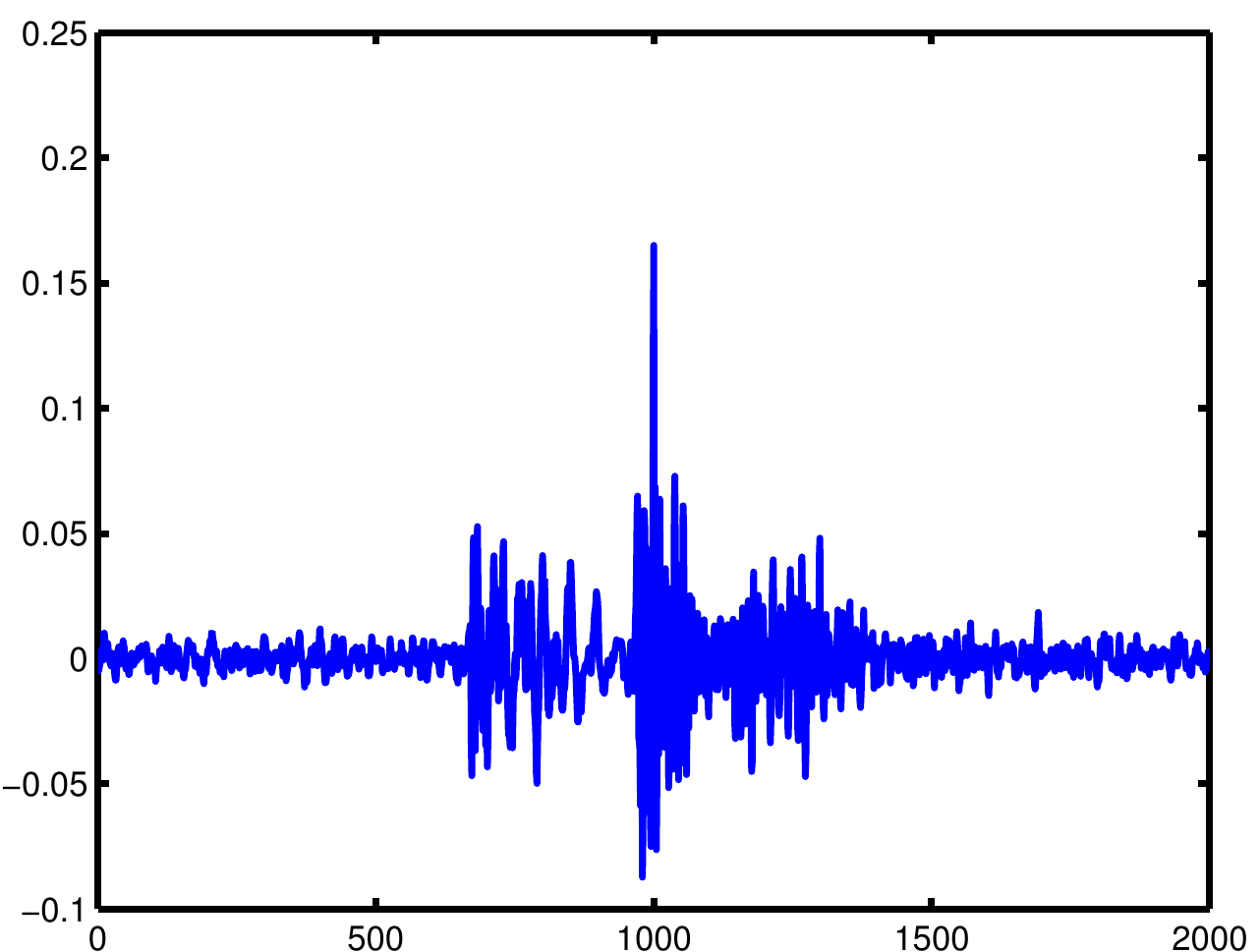} &
\includegraphics[height=5cm,width=7.5cm]{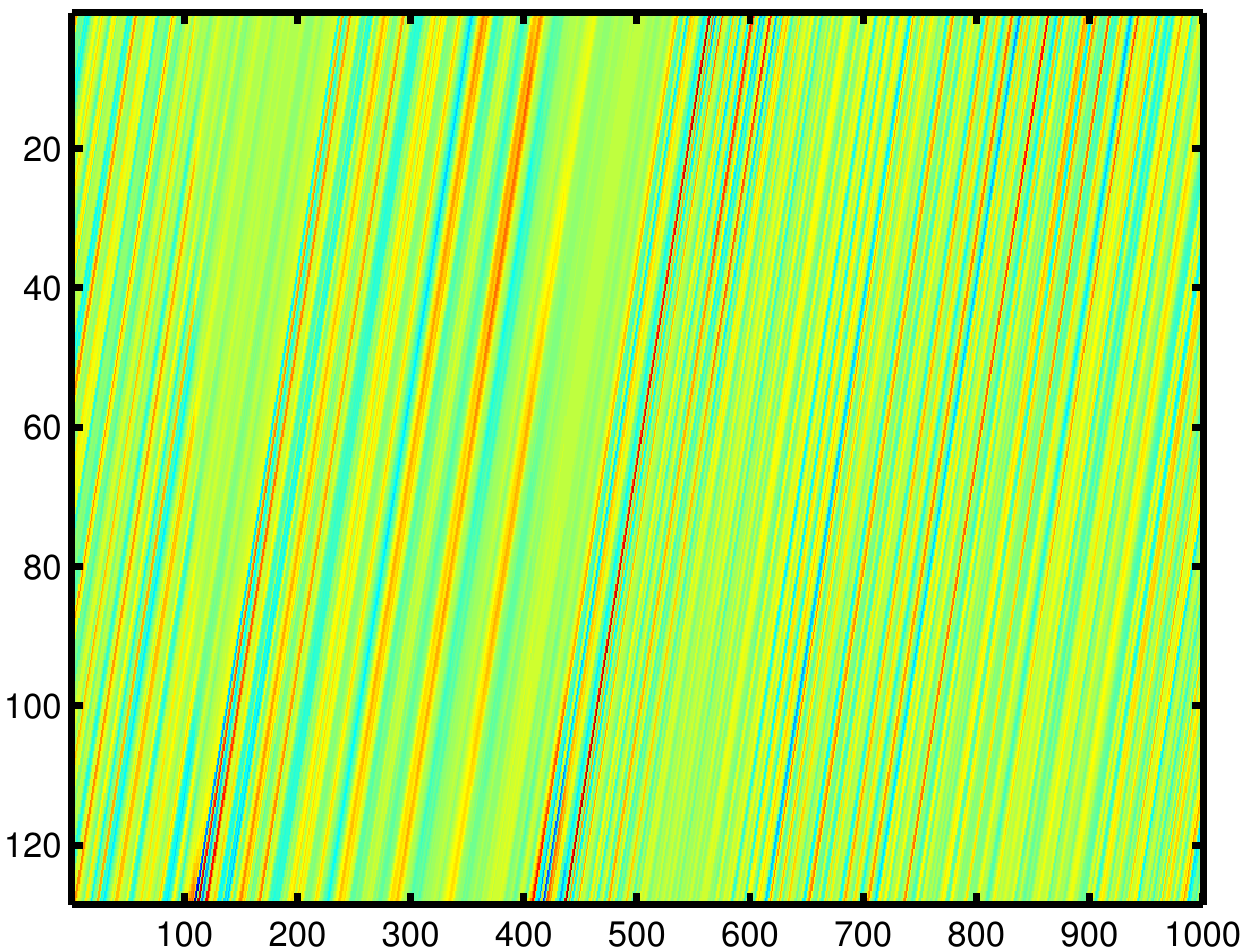}
\end{tabular}
\caption{Left: Example of detected click with $n=2000$. Right: extracted local features with $p=128$, $L=1000$ (one local feature per column).}
\label{patch_extraction}
\end{center}
\end{figure*}

As local features, we extract simply some local signal patches of $p\leq n$ samples (typically $p=128$) and denoted by $\g{z}_{i,l}^j\in\mathbb{R}^p$. Furthermore all $\g{z}_{i,l}^j$ are $\ell_2$ normalized. For each $\g{c}_{i}^j$, a total of $L$ local patches $\g{Z}_{i}^j\triangleq\{\g{z}_{i,l}^j\}$, $l=1,\ldots,L$ equally spaced of $\lceil\frac{n}{L}\rceil$ samples are retrieved (see Fig.~\ref{patch_extraction}). All local patches associated with the $j^{th}$ hydrophone is denoted by $\g{Z}^{j}\triangleq\{\g{Z}_i^{j}\}$, $i=1,\ldots,N^j$ while $\g{Z}\triangleq\{\g{Z}^j\}$ is denoting all the local patches matrix for all hydrophones. A final post-processing consists in uncorrelate local features by PCA training and projection with $p'\leq p$ dimensions.

\subsection{Local feature encoding by sparse coding}

In order to obtain a global robust representation of $\g{c}\subset\g{C}$, each associated local patch $\g{z}\subset\g{Z}$ are first linearly encoded \textit{via} the vector $\g{\alpha}\in\mathbb{R}^k$ such as $\g{z}\approx\g{D}\g{\alpha}$ where $\g{D}\triangleq [\g{d}_1,\ldots,\g{d}_k]\in\mathbb{R}^{p\times k}$ is a pre-trained dictionary matrix whose column vectors respect the constraint $\g{d}_j^T\g{d}_j=1$. In a first attempt to solve this linear problem, $\g{\alpha}$ can be the solution of the Ordinary Least Square (OLS) problem:
\begin{equation}\label{1}
l_{OLS}(\g{\alpha}|\g{z};\g{D}) \triangleq \min_{\g{\alpha} \in \mathbb{R}^k}\left\lbrace\frac{1}{2}\Vert \g{z} - \g{D}\g{\alpha}\Vert_2^2 \right\rbrace.
\end{equation}
OLS formulation can be extended to include regularization term avoiding overfitting. We obtain the ridge regression (RID) formulation:
\begin{equation}
l_{RID}(\g{\alpha}|\mathbf{z};\mathbf{D}) \triangleq \min_{\g{\alpha} \in \mathbb{R}^k}\left\lbrace\frac{1}{2}\Vert \g{z} - \g{D}\g{\alpha}\Vert_2^2 + \beta \Vert \g{\alpha} \Vert_2^2 \right\rbrace.
\end{equation}
This problem have an analytic  solution $\g{\alpha} = (\g{D}^T\g{D} + \beta\g{I}_k)^{-1}\g{D}^T\g{z}$. Thanks to semi-positivity of $\g{D}^T\g{D} + \beta\g{I}_k$, we can use a cholesky factor on this matrix to solve efficiently this linear system. In order to decrease reconstruction error and to have a sparse solution, this problem can be reformuled as a constrained Quadratic Problem (QP):
\begin{equation}
l_{SC}(\g{\alpha}|\g{z};\g{D}) \triangleq \min_{\g{\alpha} \in \mathbb{R}^k} \frac{1}{2}\Vert \g{z} - \g{D} \g{\alpha} \Vert_2^2 \ s.t. \ \ \Vert\boldsymbol{\alpha}\Vert_1 = 1.
\end{equation}
To solve this problem, we can use a QP solver involving high combinatorial computation to find the solution. Under RIP assumptions \cite{Tibshirani94regressionshrinkage}, a greedy approach can be used efficiently to solve and eq. 3 and this latter can be rewritten as:
\begin{equation}\label{2}
l_{SC}(\g{\alpha}|\g{z};\g{D}) \triangleq \min_{\g{\alpha} \in \mathbb{R}^k} \frac{1}{2}\Vert \g{z} - \g{D} \g{\alpha} \Vert_2^2 + \lambda\Vert\g{\alpha}\Vert_1,
\end{equation}
where $\lambda$ is a regularization parameter which controls the level of sparsity. This problem is also known as basis pursuit \cite{Chen98atomicdecomposition} or the Lasso \cite{Tibshirani94regressionshrinkage}. To solve this problem, we can use the popular Least angle regression (LARS) algorithm.

\subsection{Pooling local codes}

The objective of pooling \cite{Boureau10atheoretical,Feng11} is to transform the joint feature representation into a new, more usable one that preserves important information while discarding irrelevant detail.
For each click signal, we usually compute $L$ codes denoted $\g{V} \triangleq \left\lbrace\g{\alpha}_i\right\rbrace$, $i = 1,\ldots,L$.
Let define $\g{v}^{j}\in\mathbb{R}^L$, $j=1,\ldots,k$ as the $j^{th}$ row vector of $\g{V}$. It is essential to use feature pooling to map the response vector $\g{v}^{j}$ into a statistic value $f(\g{v}^{j})$ from some spatial pooling operation $f$. We use $\g{v}^{j}$, the response vector, to summarize the joint distribution of the $j^{th}$ compounds of local features over the region of interest (ROI). We will consider the $\ell_{\mu}$-norm pooling and defined by:
\begin{equation}
f_n(\g{v};\mu) = \left(\sum_{m=1}^L |v_m|^{\mu}\right)^{\frac{1}{\mu}} \ \ s.t. \ \mu \neq 0.
\end{equation}
The parameter $\mu$ determines the selection policy for locations. When $\mu = 1$, $\ell_{\mu}$-norm pooling is equivalent to sum-pooling and aggregates the responses over the entire region uniformly. When $\mu$ increases, $\ell_{\mu}$-norm pooling approaches max-pooling. We can note the value of $\mu$ tunes the pooling operation to transit from sum-pooling to max-pooling.


\subsection{Pooling codes over a temporal pyramid}
In computer vision, Spatial Pyramid Matching (SPM) is a technic (introduced by \cite{Lazebnik2006}) which improves classification accuracy by performing a more robust local analysis. We will adopt the same strategy in order to pool sparse codes over a temporal pyramid (TP) dividing each click signal into ROI of different sizes and locations. Our TP is defined by the matrix $\g{\Lambda}$ of size $(P \times 3)$ \cite{sebastienparis_xanaduhalkias_herveglotin_2013}:
\begin{equation}
\g{\Lambda} = [\g{a}, \g{b}, \g{\Omega}],
\end{equation}
where $\g{a}$, $\g{b}$, $\g{\Omega}$ are 3 $(P \times 1)$ vectors representing subdivision ratio, overlapping ratio and weights respectively. $P$ designs the number of layers in the pyramid. Each row of $\g{\Lambda}$ represents a temporal layer of the pyramid, \textit{i.e.} indicates how do divide the entire signal into sub-regions possibly overlapping. For the $i^{th}$ layer, the click signal is divided into $D_i=\lfloor\frac{1-a_i}{b_i}+1\rfloor$ ROIs where $a_i$, $b_i$ are the $i^{th}$ elements of vector $\g{a}$, $\g{b}$ respectively. For the entiere TP, we obtain a total of $D=\sum\limits_{i=1}^{P}D_i$ ROIs. Each click signal $\g{c}$ $(n \times 1)$ is divided into temporal ROI $\g{R}_{i,j}$, $i=1,\ldots,P$, $j=1,\ldots,D_i$ of size $(\lfloor a_i.n\rfloor \times 1)$. All ROIs of the $i^{th}$ layer have the same weight $\Omega_i$. For the $i^{th}$ layer, ROIs are shifted by $\lfloor b_i.n\rfloor$ samples. A TP with $\g{\Lambda} = \left[\begin{array}{ccc}
1 & 1 & 1  \\ \frac{1}{2} & \frac{1}{4} & 1  \end{array}\right]$ is designing a 2-layers pyramid with $D=1+4$ ROIs, the entiere signal for the first layer and $4$ half-windows of $\frac{n}{2}$ samples with $25\%$ of overlapping for the second layer.
At the end of pooling stage over $\g{\Lambda}$, the global feature $\g{x}\in\mathbb{R}^d$, $d=D.k$ is defined by the weighted concatenation (by factor $\Omega_i)$ of $L$ pooled codes associated with $\g{c}$.

\subsection{Dictionary learning}

\begin{figure}[!ht]
\begin{center}
\includegraphics[height=5.5cm,width=7.5cm]{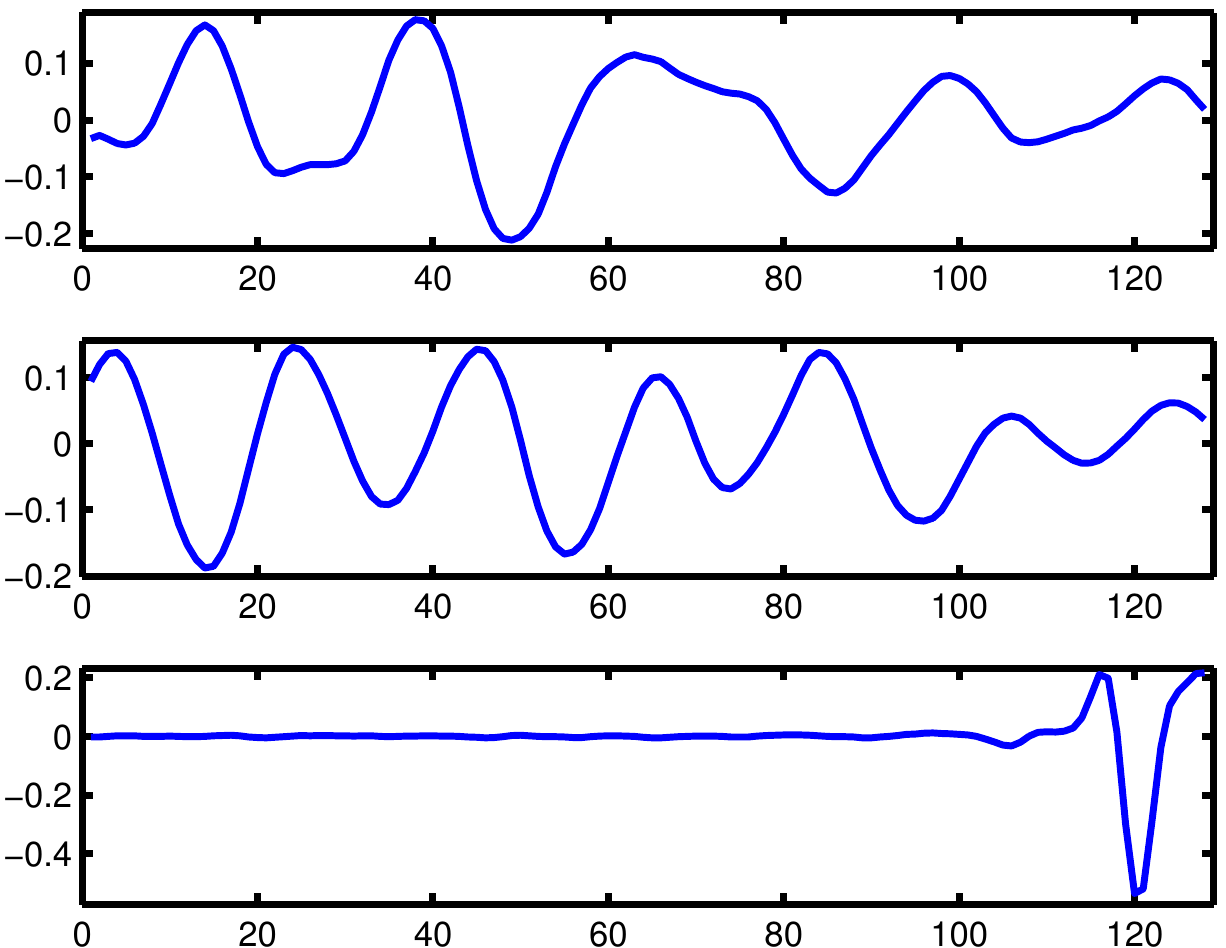}
\caption{Example of trained dictionary basis with sparse coding.}
\label{baha_click_range_dico}
\end{center}
\end{figure}

To encode each local features by sparse coding (see eq.~\ref{2}), a dictionary $\g{D}$ is trained offline with an important collection of $M\leq N.L$ local features as input. One would minimize the regularized empirical risk $\mathcal{R}_M$:
\begin{equation}
\begin{array}{c}
\mathcal{R}_M(\g{V},\g{D}) \triangleq \displaystyle\frac{1}{M}\sum\limits_{i=1}^M \frac{1}{2}\Vert \g{z}_i - \g{D} \g{\alpha}_i \Vert_2^2 + \lambda\Vert\g{\alpha}_i\Vert_1
\\
\\
\ s.t. \ \g{d}_j^T\g{d}_j=1.
\end{array}
\end{equation}
Unfortunatly, this problem is not jointly convex but can be optimized by alternating method:
\begin{equation}
\mathcal{R}_M(\g{V}|\g{\hat{D}}) \triangleq \frac{1}{M}\sum\limits_{i=1}^M \frac{1}{2}\Vert \g{z}_i - \g{\hat{D}} \g{\alpha}_i \Vert_2^2 + \lambda\Vert\g{\alpha}_i\Vert_1,
\end{equation}
which can be solved in parallel by LASSO/LARS and then:
\begin{equation}
\mathcal{R}_M(\g{D}|\g{\hat{V}}) \triangleq \frac{1}{M}\sum\limits_{i=1}^M \frac{1}{2}\Vert \g{z}_i - \g{D} \g{\hat{\alpha}}_i \Vert_2^2 \ \ s.t. \ \g{d}_j^T\g{d}_j=1.\label{eq_dico}
\end{equation}
Eq.~\ref{eq_dico} have an analytic solution involving a large matrix $(k\times k)$ inversion and a large memory occupation for storing the matrix $\g{V}$ $(k\times M)$. Since $M$ is potentially very large (up to 1 million), an online method to update dictionary learning is prefered \cite{Mairal_2009}. Figure \ref{baha_click_range_dico} depicts 3 dictionary basis vectors learned \textit{via} sparse coding. As depicted, some elements reprensents more impulsive responses while some more harmonic responses.

\section{Range and azimuth logistic regression from global features}

After the pooling stage, we extracted unsupervisly $N$ global features $\g{X}\triangleq\{x_i\}\in\mathbb{R}^{d\times N}$. We propose to regress \textit{via} logistic regression both range $r$ and azimuth $az$ (in $x-y$ plan, when animal reach surface to breath) from the animal trajectory groundtruth denoted $\g{y}$. For the current train/test splitsets of the data, such as $\g{X}=\g{X}_{train}\bigcup\g{X}_{test}$, $\g{y}=\g{y}_{train}\bigcup\g{y}_{test}$ and $N=N_{train}+N_{test}$, $\forall$ $\{\g{x}_i,y_i\}\in\g{X}_{train}\times \g{y}_{train}$, we minimize:
\begin{equation}
\widehat{\g{w}}_{\theta}=\arg\min\limits_{\g{w}_{\theta}}\left\{\frac{1}{2}\g{w}_{\theta}^T\g{w}_{\theta} + C\sum\limits_{i=1}^{N_{train}}\log(1+e^{-y_i\g{w}_{\theta}^T\g{x}_i})\right\},\label{logistic_regression}
\end{equation}
where $y_i$ denotes $r_i$ and $az_i$ for $\theta=r$ and $\theta=az$ respectively. Eq.~\ref{logistic_regression} can be efficiently solved for example with Liblinear software \cite{liblinear2008}. In the test part, range and azimuth for any $\g{x}_i\in\g{X}_{test}$ are recontructed linearly by $\widehat{r}_i=\widehat{\g{w}}_r^T\g{x}_i$ and by $\widehat{az}_i=\widehat{\g{w}}_{az}^T\g{x}_i$ respectively.

\section{Experimental results}

\subsection{bahamas2 dataset}

This dataset \cite{3Dtracking} contains a total of $N=6134$ detected clicks for $H=5$ different hydrophones (named $H^7$, $H^8$, $H^9$, $H^{10}$ and $H^{11}$ and with $N^7=1205$, $N^8=1238$, $N^9=1241$, $N^{10}=1261$ and $N^{11}=1189$ respectively).

\begin{figure}[!ht]
\begin{center}
\includegraphics[height=5.5cm,width=7.5cm]{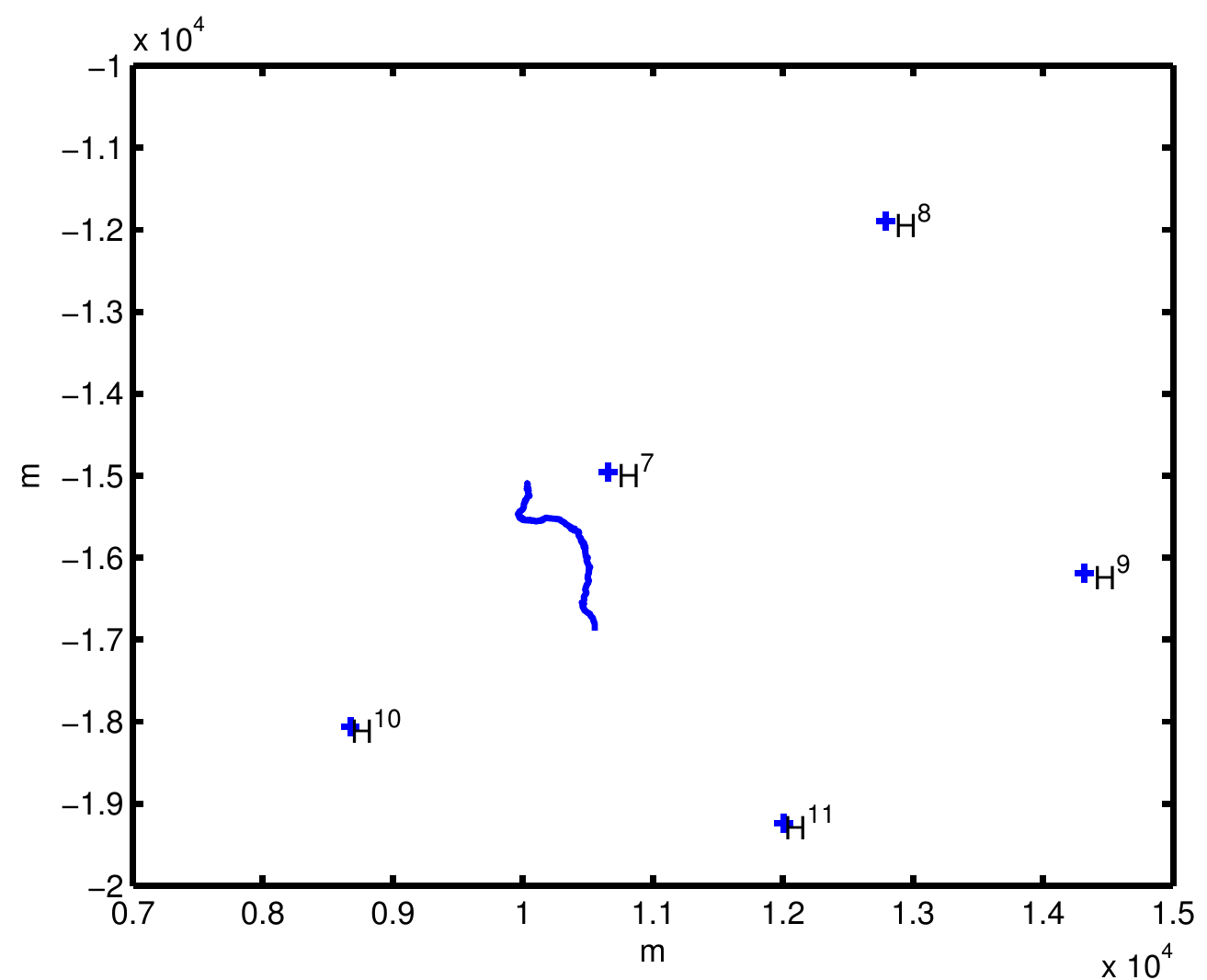}
\caption{The 2D trajectory (in $x−y$ plan) of the single sperm whale observed during $25$ min and corresponding hydrophone’s positions.}
\label{true_trajectory}
\end{center}
\end{figure}
To extract local features, we chose $n=2000$, $p=128$ and $L=1000$ (tuned by model selection). For both the dictionary learning and the local features encoding, we chose $\lambda=0.2$ and fixed $15$ iterations to train dictionary on a subset of $M=400.000$ local features drawn uniformaly. We performed
$K=10$ cross-validation where training sets reprensented $70\%$ of the total of extracted global features, the rest for the testing sets. Logistic regression parameter $C$ is tuned by model selection. We compute the average root mean square error (ARMSE) of range/azimuth estimates per hydrophone: $ARMSE(l)=\frac{1}{K}\sum\limits_{i=1}^{K}\sqrt{\sum\limits_{j=1}^{N_{test}^l}(y_{i,j}^l-\widehat{y}_{i,j}^l)^2}$ where $y_{i,j}^l$, $\widehat{y}_{i,j}^l$
 and $N_{test}^l$ represent the ground truth, the estimate and the number of test samples for the $l^{th}$ hydrophone respectively. The global ARMSE is then calculated by $\overline{ARMSE}=\frac{1}{H}\sum\limits_{l=1}^{H}ARMSE(l)$.

 \subsection{$\ell_{\mu}$-norm pooling case study}

 For prilimary results, we investigate the influence of the $\mu$ parameter during the pooling stage. We fix the number of dictionary basis to $k=128$ and the temporal pyramid equal to $\g{\Lambda}_1=\left[1,1,1\right]$, \textit{i.e.} we pool sparse codes on whole the temporal click signal.
\begin{figure}[!ht]
\begin{center}
\includegraphics[height=5.5cm,width=7.5cm]{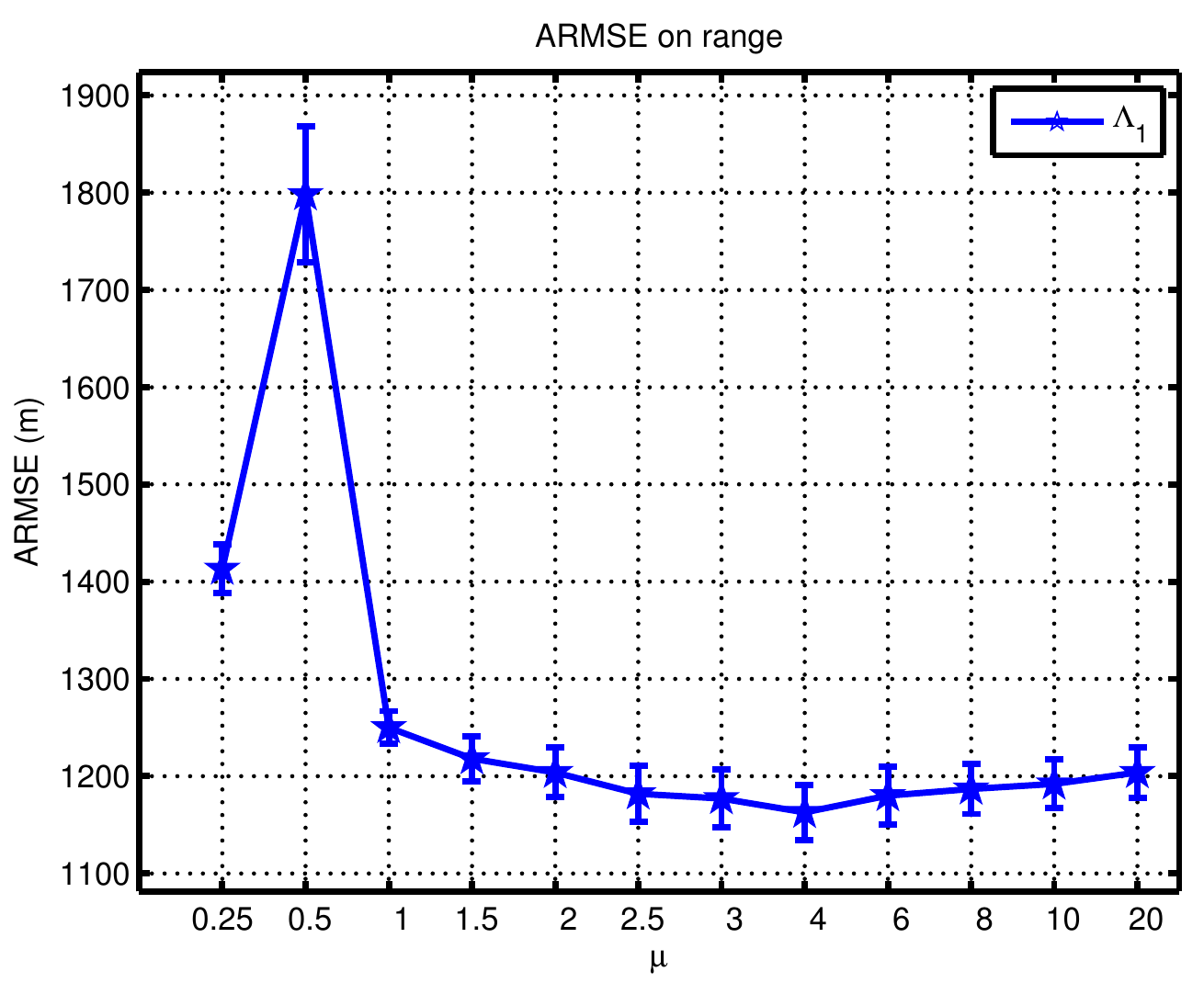}
\caption{$\overline{ARMSE}$ vs. $\mu$ for range estimation.}
\label{baha_click_range_nu_pooling}
\end{center}
\end{figure}
A value of $\mu=\{3,4\}$ seems to be a good choice for this pooling procedure. For $\mu\geq20$, results are similar to those obtained by max-pooling. For azimuth, we observe also the same range of $\mu$ values.

 \subsection{Range and azimuth regression results}

 Here, we fixed the value of $\mu=3$ and we varied the number of dictionary basis $k$ from $128$ to $4096$ elements. We also investigated the influence of the temporal pyramid and we give results for two particulary choices: $\g{\Lambda}_1=\left[1,1,1\right]$ and $\g{\Lambda}_2=\small\left[\begin{array}{ccc}1 & 1 &1\\ \frac{1}{3} & \frac{1}{3}&1\end{array}\right]$. For $\g{\Lambda}_2$, the sparse are first pooled over all the signal then pooled over 3 non-overlapping windows for a total of $1+3=4$ ROIs.
 In order to compare results of our presented method, we also give results for an hand-craft feature \cite{IFA_DCL} specialized for spermwhales and based on the spectrum of the most energetic pulse détected inside the click. This specialized feature, denoted \textit{Spectrum feature}, is a 128 points vector.

\begin{figure}[!ht]
\begin{center}
\includegraphics[height=5.5cm,width=7.5cm]{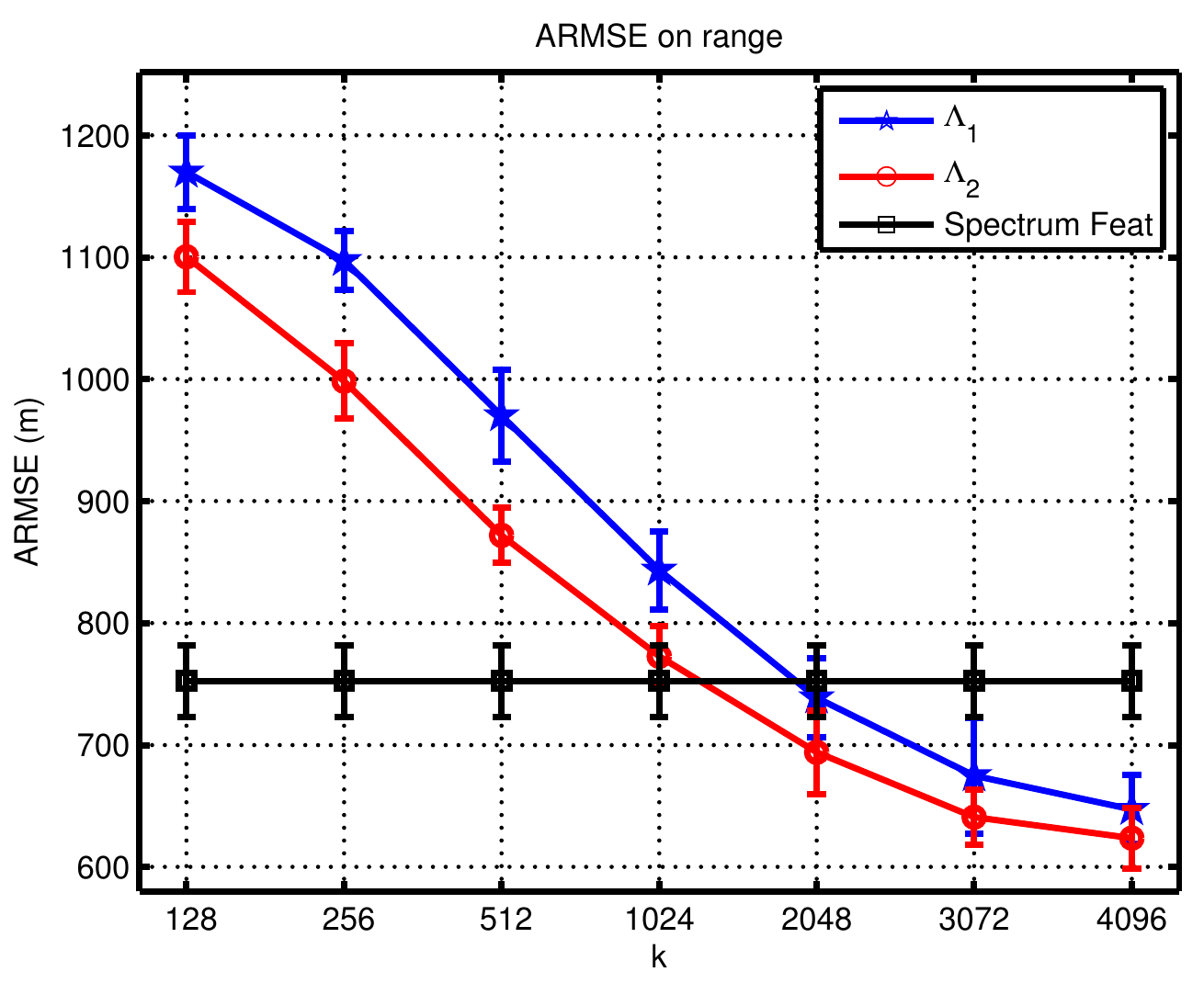}
\caption{$\overline{ARMSE}$ vs. $k$ for range estimation with $\mu = 3$.}
\label{baha_click_range}
\end{center}
\end{figure}

\begin{figure}[!ht]
\begin{center}
\includegraphics[height=5.5cm,width=7.5cm]{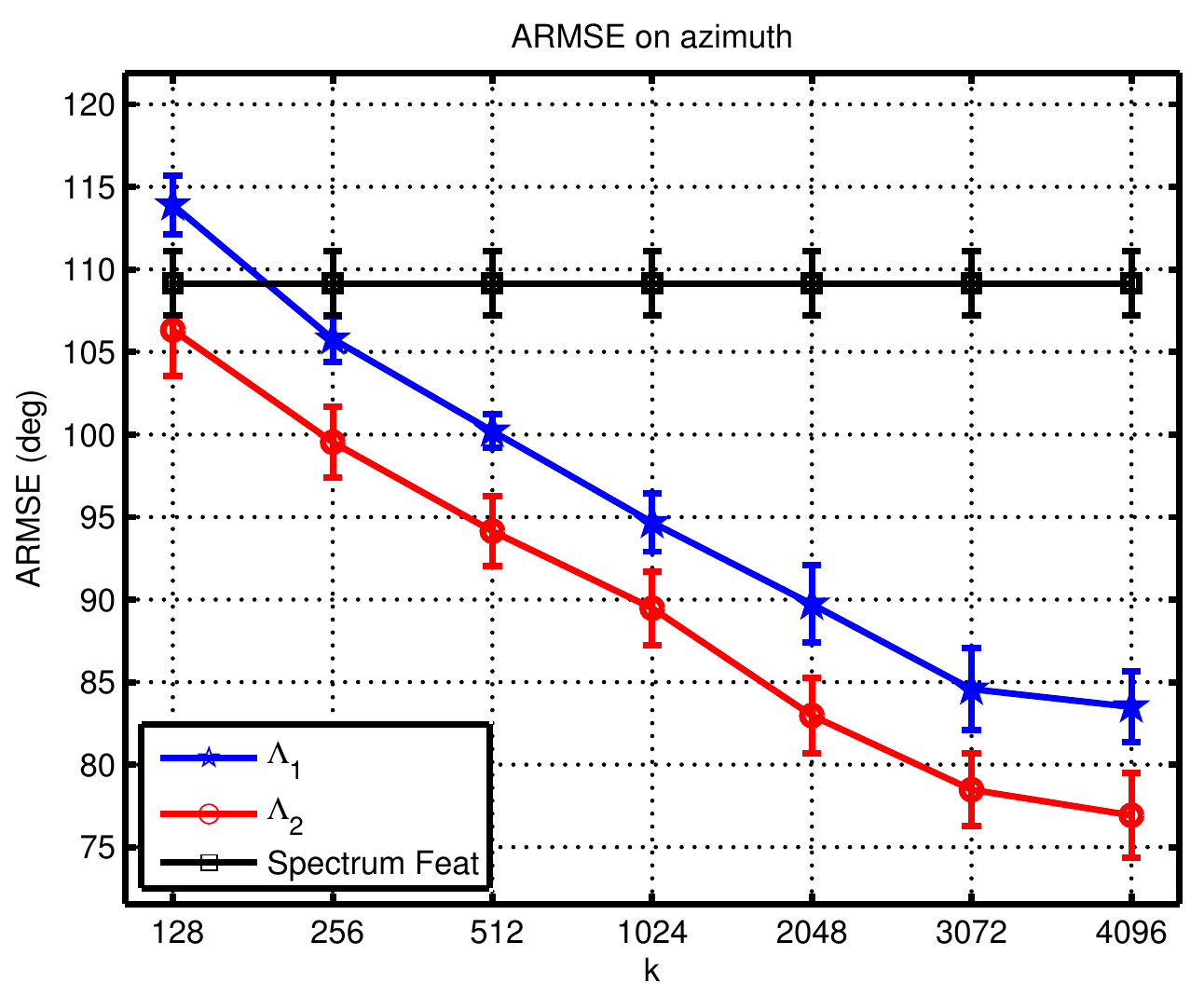}
\caption{$\overline{ARMSE}$ vs. $k$ for azimuth estimation with $\mu = 3$.}
\label{baha_click_azimuth}
\end{center}
\end{figure}

For both range and azimuth estimate, from $k=2048$, our method outperforms results of the \textit{Spectrum feature} and particulary for azimuth estimate. Using a temporal pyramid for pooling permits also to improve slightly results.

\section{Conclusions and perspectives}

We introduced in the paper, for spermwhale localization, a BoF approach \textit{via} sparse coding delivering rough estimates of range and azimuth of the animal, specificaly towarded for mono-hydrophone configuration. Our proposed method works directly on the click signal without any prior pulses detection/analysis while being robust to signal transformation issue by the propagation. Coupled with non-linear filtering such as particle filtering \cite{Arulampalam02}, accurate animal position estimation could be perform even in mono-hydrophone configuration. Applications for anti-collision system and whale whatching are targeted with this work.

As perspective, we plan to investigate other local features such as spectral features, MFCC \cite{Davis80,Rabiner93}, Scattering transform features \cite{AndenM11}. These latter can be considered as a hand-craft first layer of a deep learning architecture with 2 layers.

\bibliography{ICML13_WS_biblio}
\bibliographystyle{icml2013}

\end{document}